\documentclass[journal]{IEEEtran}
\usepackage{amsmath,amsfonts}
\usepackage{algorithmic}
\usepackage{algorithm}
\usepackage{array}
\usepackage[caption=false,font=normalsize,labelfont=sf,textfont=sf]{subfig}
\usepackage{textcomp}
\usepackage{stfloats}
\usepackage{url}
\usepackage{verbatim}
\usepackage{graphicx}
\usepackage{cite}
\usepackage{booktabs}
\usepackage{multirow}
\usepackage{amssymb}
\usepackage{xcolor}

\begin{document}

\title{BookNet: Dual-Page Book Image Rectification via Cross-Page Attention}

\author{
 Shaokai Liu,~\IEEEmembership{Member,~IEEE}, 
 Hao Feng*,
 Bozhi Luan, 
 Min Hou,
 \par
 Jiajun Deng,~\IEEEmembership{Member,~IEEE}, and~Wengang Zhou*,~\IEEEmembership{Senior Member,~IEEE}

\IEEEcompsocitemizethanks{
\IEEEcompsocthanksitem The computations were performed on the HPC Platform of Hefei University of Technology and the GPU cluster built by the MCC Lab of the School of Information Science and Technology, USTC.
\IEEEcompsocthanksitem This work was supported in part by the National Natural Science Foundation of China under Contract 82604079, 62133004, and 62322308, in part by the Fundamental Research Funds for the Central Universities (Grant No. JZ2025HGQA0641 and JZ2026HGTS0202), and in part by the Scientific Research Innovation Capability Support Project for Young Faculty (Grant No. SRICSPYF-ZY2025006).
\IEEEcompsocthanksitem Shaokai Liu is with the School of Management and Intelligence Science and the Key Laboratory of Process Optimization and Intelligent Decision-Making, Ministry of Education, Hefei University of Technology, Hefei 230009, China.
E-mail: liushaokai@hfut.edu.cn
\IEEEcompsocthanksitem Min Hou is with the School of Computer Science and Information Engineering and the Key Laboratory of Knowledge Engineering with Big Data, Ministry of Education, Hefei University of Technology, Hefei 230009, China.
E-mail: hmhoumin@gmail.com
\IEEEcompsocthanksitem Hao Feng, Bozhi Luan, Jiajun Deng and Wengang Zhou are with University of Science and Technology of China, Hefei, 230027, China.
E-mail: \{haof, lbz0075\}@mail.ustc.edu.cn; \{dengjj,zhwg\}@ustc.edu.cn
\IEEEcompsocthanksitem *Corresponding authors: Hao Feng and Wengang Zhou.
 
}

}

\maketitle

\begin{abstract}
Book image rectification presents unique challenges due to its complex geometric distortions from binding constraints, where left and right pages exhibit distinctly asymmetric curvature patterns. However, existing single-page document image rectification methods fail to capture the coupled geometric relationships between adjacent pages in books. In this work, we introduce BookNet, the first end-to-end deep learning framework specifically designed for dual-page book image rectification. BookNet adopts a dual-branch architecture with cross-page attention mechanisms, enabling it to estimate warping flows for both individual pages and the book spread, explicitly modeling how left and right pages influence each other. Moreover, to address the absence of specialized datasets, we present Book3D, a large-scale synthetic dataset for training, and Book100, a comprehensive real-world benchmark for evaluation. Extensive experiments demonstrate that BookNet outperforms existing state-of-the-art methods on book image rectification. Code and dataset are available at \url{https://github.com/Liu-Shaokai/BookNet}.

\end{abstract}


\section{Introduction}
\IEEEPARstart{C}{amera-captured} document images have become increasingly prevalent in digital workflows, offering greater convenience and accessibility than traditional scanning. However, these images often suffer from geometric distortions, especially for bound documents such as books. Unlike single-page document dewarping, book image rectification presents unique challenges due to binding constraints that create asymmetric deformations across left and right pages. Rectification is critical for downstream applications including cultural heritage digitization~\cite{yang2025predicting}, knowledge management~\cite{zenouz2021knowledge,xu2025docksrag}, and multimodal understanding~\cite{li2024monkey,feng2024docpedia}.

\begin{figure}[!t]
\centering
\includegraphics[width=\linewidth]{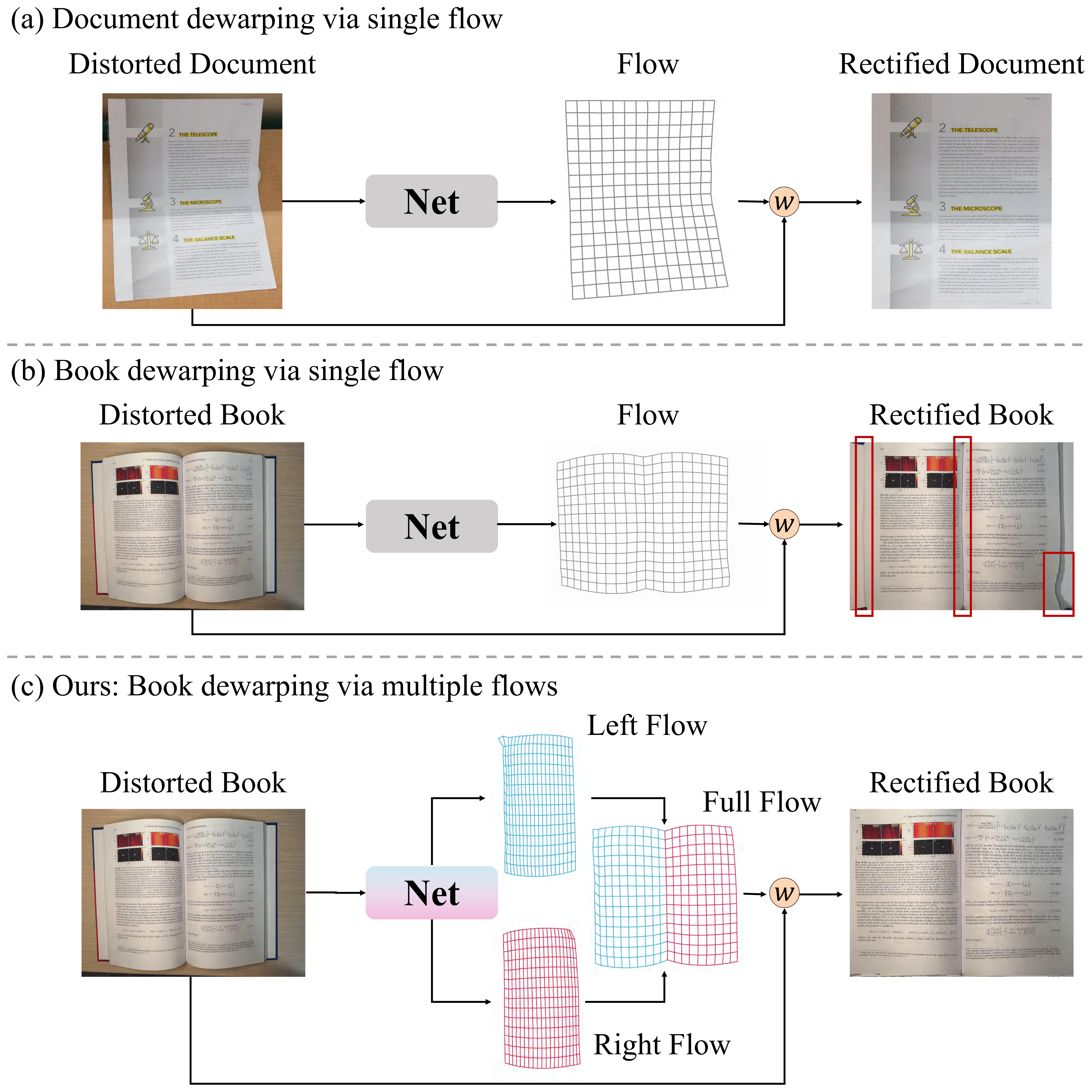}
\caption{Rectification paradigm comparison. (a) Conventional single flow for individual pages. (b) Single flow fails on books. (c) Our multi-flow solution effectively rectifies books by predicting separate flows (left, right, full).}
\label{fig:paradigm_comparison}
\end{figure}

While industrial solutions such as flatbed scanners or specialized overhead cameras can partially mitigate these distortions, they require expensive equipment and controlled environments, limiting their accessibility. Early computational approaches relied on specialized hardware~\cite{brown2001document,zhang2008improved,meng2014active}, shape-from-shading~\cite{wada1997shape}, 3D reconstruction~\cite{pal20133d}, or model-based methods~\cite{wu2007model,liang2008geometric,meng2012metric,he2013book} with explicit geometric modeling. However, these methods often required specific capture conditions, manual intervention, or high computational cost, limiting their deployment in real-world scenarios.

The advent of deep learning has revolutionized document image rectification, enabling end-to-end learning of complex deformation patterns through data-driven warping flow prediction. However, existing deep learning methods~\cite{ma2018docunet,das2019dewarpnet,feng2021doctr,das2021end,feng2022geometric,jiang2022revisiting,xue2022fourier,ma2022learning,zhang2022marior,li2023layout,feng2025docscanner,feng2023doctrplus,han2025docmamba} primarily target single-page documents. As shown in Fig.~\ref{fig:paradigm_comparison}(a), these methods employ a single warping flow to rectify individual pages. When applied to books, however, this approach fails to capture the asymmetric deformations where left and right pages exhibit distinct patterns influenced by binding constraints (Fig.~\ref{fig:paradigm_comparison}(b)). A straightforward alternative of applying single-page methods separately followed by stitching also proves problematic, as it requires either precise spine centering for industrial solutions or dual captures with manual alignment for learning-based methods. Moreover, such stitching risks introducing boundary artifacts, text discontinuities, and misalignment between pages.

To address these limitations, we propose BookNet, the first end-to-end deep learning framework specifically designed for dual-page book image rectification. Our key insight is that effective book rectification requires modeling both page-specific and cross-page deformation patterns. As shown in Fig.~\ref{fig:paradigm_comparison}(c), our approach predicts three complementary warping flows: left flow for the left page, right flow for the right page, and full flow for the book spread. The page-specific flows capture distinct deformation characteristics of individual pages, while the full flow provides holistic rectification guidance by modeling their interactions across the binding region.

To facilitate comprehensive evaluation in this underexplored area, we contribute specialized datasets tailored for book image rectification. We construct Book3D, a large-scale synthetic training dataset containing 56,000 high-resolution book images with realistic 3D deformation patterns derived from academic papers. This dataset effectively enables training of robust book image rectification models. Additionally, we introduce Book100, a real-world evaluation benchmark comprising 100 diverse book images captured under various conditions. Each image is paired with high-quality reference scans, providing comprehensive assessment capabilities for real-world scenarios. This dataset addresses the critical gap in book-specific evaluation resources and establishes a standardized benchmark for fair comparison of rectification models.

In summary, our main contributions are:
\begin{itemize}
    \item We propose BookNet, the first end-to-end deep learning framework
    specifically designed for dual-page book image rectification, adopting a
    cross-page attention architecture with specialized dual branches for modeling
    asymmetric page deformations and inter-page dependencies.
    \item We construct a training dataset and an evaluation benchmark specifically designed for book image rectification. The training dataset uses synthetic rendering and contains 56,000 samples. The evaluation benchmark contains 100 real book photos with their corresponding scans.
    \item We conduct extensive experiments demonstrating that our approach significantly outperforms state-of-the-art methods, achieving superior performance on real-world book images with substantial improvements in geometric accuracy and content preservation.
\end{itemize}

\begin{figure*}[!t]
  \centering
  \includegraphics[width=\linewidth]{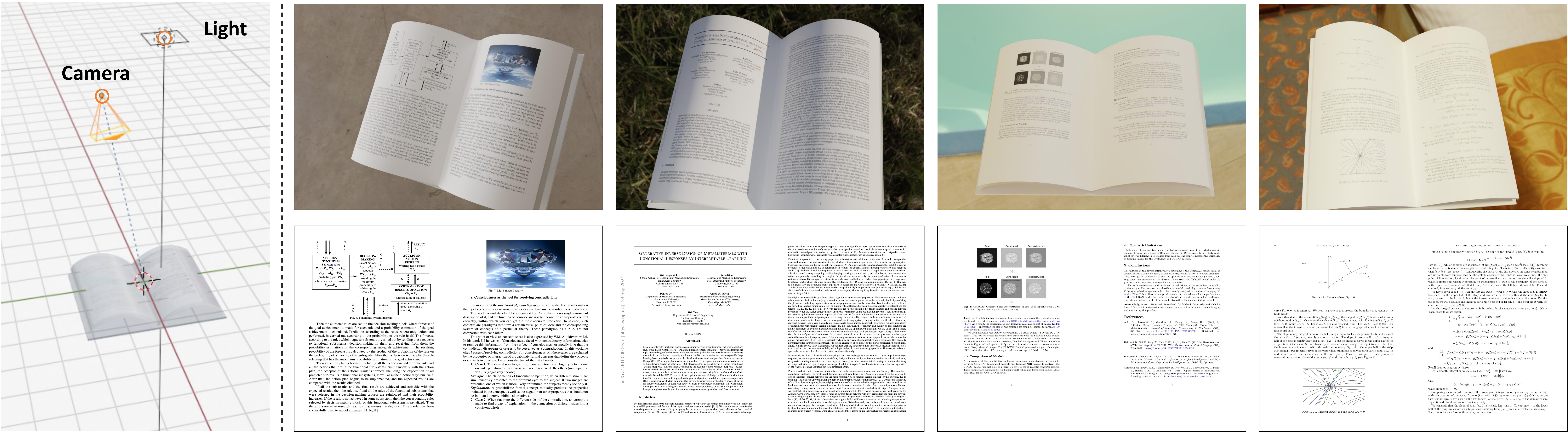}
  \caption{Book3D synthetic dataset generation pipeline and representative samples. Left: Blender rendering workspace showcasing the 3D book modeling environment with randomized book proportions and camera poses. Right: Rendered book samples from diverse arXiv academic papers, demonstrating realistic geometric deformations under varied illumination conditions and viewing angles. Top row shows the rendered synthetic book images, while bottom row displays the corresponding ground truth arXiv paper images.}
  \label{fig:book3d_samples}
\end{figure*}

\section{Related Work}
\label{sec:Related}

In this section, we classify prior work on book and document image rectification into two broad categories: traditional book and document rectification methods and deep learning-based approaches. We then discuss their limitations in handling book-specific challenges such as dual-page structures and binding-induced deformations.

\subsection{Traditional Book and Document Rectification}

Traditional methods rely on specialized hardware or geometric modeling, but are limited by equipment requirements and feature detection reliability.

\subsubsection{Hardware-Based Approaches}
Early book and document digitization research utilized specialized hardware for complex 3D geometry reconstruction. Wada et al.~\cite{wada1997shape} pioneered shape-from-shading techniques using interreflections in flatbed scanners. Brown and Seales~\cite{brown2001document} integrated structured light systems with physical document models for manuscript restoration. Subsequent works~\cite{zhang2008improved,meng2014active} employed laser scanning and structured beam illumination for heritage preservation. Galarza et al.~\cite{galarza2018time} combined Time-of-Flight sensors with cameras for assistive reading applications. However, these hardware-dependent methods require specialized equipment, limiting practical deployment.

\subsubsection{Multi-View and Geometric Model-Based Methods}
To reduce hardware dependency, researchers explored alternative approaches. Multi-view methods~\cite{yamashita2004shape,koo2009composition} employed stereo vision for book reconstruction. Subsequent work extended this paradigm to more diverse scenarios, with Kim et al.~\cite{kim2013dewarping} adapting it for mobile phone cameras using structure-from-motion, and You et al.~\cite{you2018multiview} extending it to handle heavily folded documents from hand-held cameras. Alternatively, geometric modeling approaches~\cite{wu2007model,liang2008geometric,meng2012metric} estimated 3D shapes from texture flow and geometric constraints without specialized hardware. Cao et al.~\cite{cao2003cylindrical} proposed a cylindrical surface model specifically for bound documents, using text line baselines as geometric cues to estimate page curvature and perform rectification through coordinate transformations. These approaches can be further divided into methods based on text lines~\cite{tian2011rectification,he2013book} and vanishing points~\cite{simon2021generic} for handling complex layouts. However, multi-view methods require multiple captures, and geometric approaches depend on reliable feature detection, both limiting robustness in challenging scenarios.

\begin{figure*}[!t]
  \centering
  \includegraphics[width=\linewidth]{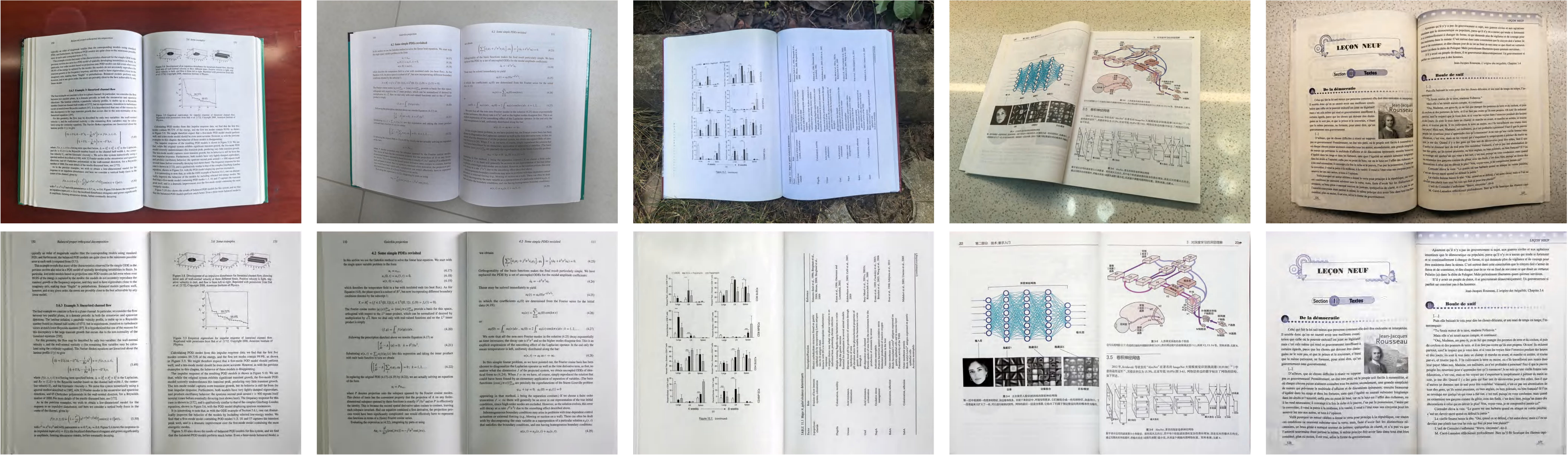}
  \caption{Representative samples from the Book100 benchmark dataset illustrating diverse capture conditions and content types. Top row: Distorted book images captured under various real-world conditions exhibiting different deformation patterns, lighting variations, and viewing angles. Bottom row: Corresponding high-quality reference scans obtained using professional overhead document cameras, providing ground truth for evaluation.}
  \label{fig:book100_samples}
\end{figure*}

\subsection{Deep Learning-Based Document Rectification}

Deep learning methods enable data-driven rectification without hand-crafted features, but existing approaches focus solely on single-page documents.

\subsubsection{Pioneering Neural Approaches}
Deep learning enabled data-driven learning of complex deformation patterns. DocUNet~\cite{ma2018docunet} pioneered pixel-wise coordinate regression using stacked U-Net~\cite{ronneberger2015u} to predict dense displacement fields. DewarpNet~\cite{das2019dewarpnet} incorporated 3D shape priors, decomposing rectification into shape estimation and texture unwarping stages trained on the large-scale Doc3D dataset. Xie et al.~\cite{xie2020dewarping} estimated pixel-wise displacements via fully convolutional networks, employing local smooth constraints for rectification.

\subsubsection{Advanced Architectures and Attention Mechanisms}
Recent innovations focused on capturing long-range spatial dependencies~\cite{ViT} crucial for deformation modeling. DocTr~\cite{feng2021doctr} introduced Transformer~\cite{Vaswani2017AttentionIA} architectures with self-attention to model global patterns. Subsequent works explored various strategies: piece-wise unwarping~\cite{das2021end}, control points-based approaches~\cite{xie2021document}, grid regularization~\cite{jiang2022revisiting}, geometric cue integration~\cite{feng2022geometric}, and Fourier-based restoration~\cite{xue2022fourier}. Building upon these advances, recent developments include learning from wild documents~\cite{ma2022learning}, margin-aware rectification~\cite{zhang2022marior}, progressive learning~\cite{feng2025docscanner}, and foreground-aware methods~\cite{li2023foreground}. DocTr++~\cite{feng2023doctrplus} extended Transformers to unrestricted documents with hierarchical structures. Additionally, diffusion-based generative models~\cite{ho2020denoising} have emerged for document restoration. Kumari and Das~\cite{kumari2025document} introduced Conditional Stable Diffusion Transformers~\cite{peebles2023scalable} for rectification, while Zhang et al.~\cite{zhang2025dvd} proposed a coordinates-based diffusion model for document dewarping. More recently, Zhao et al.~\cite{zhao2025uni} developed Uni-DocDiff, a unified document restoration model based on diffusion that handles multiple degradation types including dewarping, though it processes single-page documents rather than dual-page books. Structural matching approaches~\cite{hertlein2025docmatcher} have also been proposed to leverage textual and structural line correspondences for dewarping. Zhou et al.~\cite{zhou2026afh} proposed adaptive feature harmonization with shuffle-based attention for robust single-page dewarping.

\subsubsection{Limitations and Our Approach}
However, a common limitation across existing methods is their singular focus on single-page documents, which overlooks book-specific challenges such as coupled page deformation, binding distortions, and asymmetric patterns. To address these challenges, we introduce BookNet, the first end-to-end deep learning framework specifically designed for book image rectification, which employs dual-branch processing and cross-page attention.

\section{Dataset}
\label{sec:dataset}

The absence of specialized datasets has been a critical bottleneck in advancing book image rectification research. To address this gap, we introduce Book3D and Book100, the first large-scale synthetic dataset and real-world benchmark designed for dual-page book image rectification.

\subsection{Book3D Synthetic Training Dataset}

Existing document image rectification training datasets, such as Doc3D~\cite{das2019dewarpnet}, focus exclusively on single-page documents, failing to capture the unique characteristics of bound volumes with asymmetric left-right page deformations and coupled geometric relationships. We construct Book3D using a physically-grounded rendering pipeline that synthesizes authentic content on a realistic 3D book mesh.

Our dataset encompasses 56,000 high-resolution book images spanning eight domains to ensure diverse visual styles and content patterns: Computer Science, Economics, Electrical Engineering, Mathematics, Physics, Quantitative Biology, Quantitative Finance, and Statistics. Source materials are derived from arXiv academic papers, preserving authentic typography and mathematical notation.

Dense flow annotations are infeasible to acquire from real books, so we follow the common practice of learning from synthetic renderings~\cite{das2019dewarpnet,zhang2026docaligner}, with particular attention to the physical realism of the rendered samples. The rendering pipeline incorporates physically-realistic geometric properties including natural page curvature from binding constraints and asymmetric deformation patterns.

We employ Blender's Cycles rendering engine. Each scene is lit either by an indoor or outdoor HDR environment map, or by randomized light sources with varying color temperature. Each sample is rendered from a randomly sampled camera pose, covering diverse viewing angles and distances rather than a fixed frontal view. Each sample includes the rendered book image and its binary segmentation mask at 1200$\times$800 resolution, together with the dense warping flow of each page. As illustrated in Fig.~\ref{fig:book3d_samples}, our rendering pipeline generates diverse book samples with realistic deformations across multiple domains. All our evaluations are conducted on Book100, an entirely real benchmark discussed next, so the reported results directly measure synthetic-to-real transfer.

\subsection{Book100 Real-World Evaluation Benchmark}

\begin{table}[!t]
\centering
\caption{Book100 Dataset Diversity Statistics}
\label{tab:book100_statistics}
\renewcommand{\arraystretch}{1.5}  
\begin{tabular}{m{2cm}|p{5cm}}
\hline
\textbf{Attribute} & \textbf{Coverage} \\
\hline
Language & English, Chinese, German, French, Japanese, and others \\
\hline
Subject & Mathematics, Physics, Management, Engineering, Computer Science, Medicine, Biology, Humanities, History, and others \\
\hline
Content Type & Text, Figures, Equations, Tables, Code, Mixed \\
\hline
Capture Time & Morning, Afternoon, Evening \\
\hline
Environment & Indoor, Outdoor \\
\hline
Lighting & Natural, Artificial, Mixed \\
\hline
Perspective & Frontal, Left-side, Right-side, Overhead, Low-angle \\
\hline
\end{tabular}
\end{table}

We contribute Book100, comprising 100 high-resolution book images captured with smartphone cameras paired with corresponding reference scans. This benchmark evaluates algorithm performance across diverse real-world scenarios encountered when digitizing books with consumer-grade devices. The dataset exhibits comprehensive diversity across multiple dimensions, as detailed in Table~\ref{tab:book100_statistics}. The collection encompasses multiple languages, diverse content categories, and varying layout complexities to ensure robust evaluation across different document characteristics.

Distorted images were captured using a Huawei Mate 60 Pro+ smartphone under diverse real-world conditions including varied lighting configurations, indoor and outdoor conditions, and multiple viewing angles typical of handheld device usage. Reference images were obtained using a professional overhead document scanner (Deli 15163 Document Camera) under controlled scanning conditions with high-resolution capture. Each reference scan undergoes manual quality verification. Representative samples from our Book100 dataset are shown in Fig.~\ref{fig:book100_samples}, demonstrating the diversity of capture conditions, content types and languages. Such diversity enables rigorous evaluation across real-world scenarios. The Book100 benchmark represents the first standardized real-world evaluation dataset specifically designed for the book image rectification task.

\begin{figure*}[!t]
  \centering
  \includegraphics[width=\linewidth]{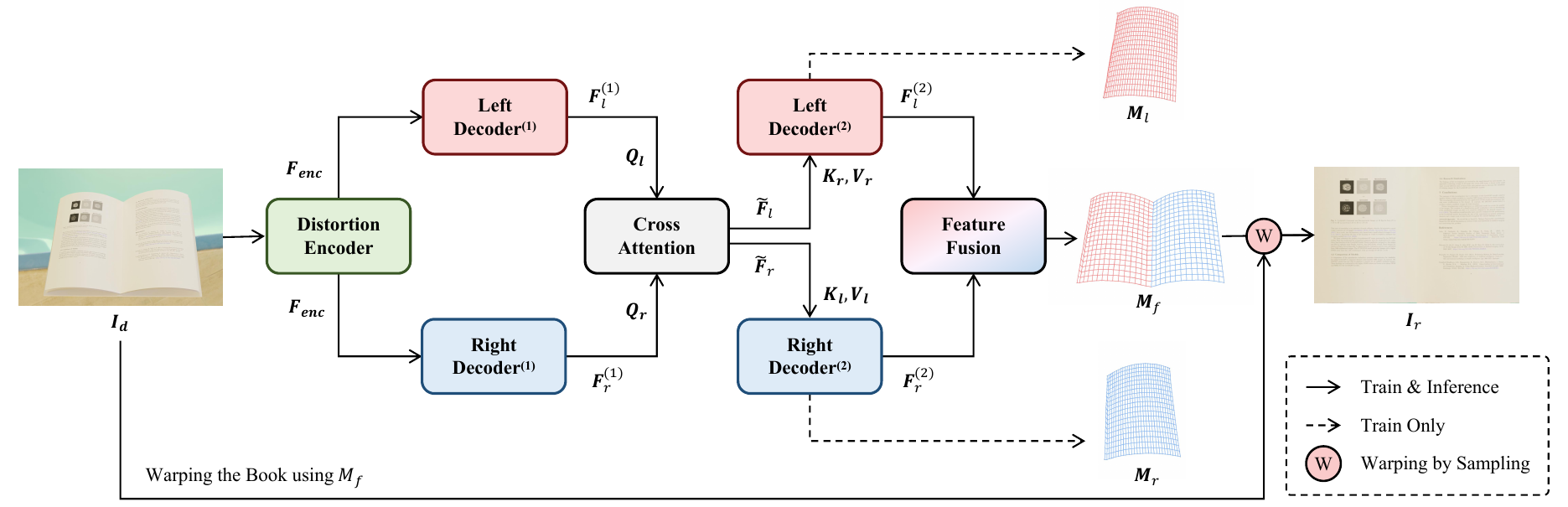}
  \caption{Overview of the proposed BookNet architecture. Given a distorted book image $\mathbf{I}_d$ containing both left and right pages, our method extracts features through a CNN backbone and Transformer encoder. The dual-branch decoder employs a two-stage architecture with cross-page attention mechanisms to process learned queries, generating the left flow $\mathbf{M}_l$, the right flow $\mathbf{M}_r$, and the full flow $\mathbf{M}_f$ for the left page, the right page, and the book spread respectively. During training, all three flows are supervised, while inference uses the full flow $\mathbf{M}_f$ for final rectification.}
  \label{fig:architecture}
\end{figure*}

\section{Method}
\label{sec:method}

An overview of the proposed BookNet is presented in Fig.~\ref{fig:architecture}. Given a distorted book image $\mathbf{I}_d \in \mathbb{R}^{H \times W \times 3}$ containing both left and right pages, we aim to estimate a rectified image $\mathbf{I}_r \in \mathbb{R}^{H \times W \times 3}$ where the geometric distortions are corrected while preserving content quality. A warping flow defined over an $H \times W$ output has dimension $H \times W \times 2$, with the last dimension holding the $(x, y)$ coordinates in the distorted input. Our BookNet predicts three such flows: $\mathbf{M}_l$ and $\mathbf{M}_r \in \mathbb{R}^{H \times \frac{W}{2} \times 2}$ for the left and right pages, and $\mathbf{M}_f \in \mathbb{R}^{H \times W \times 2}$ for the book spread. During training, all three warping flows are supervised to ensure comprehensive geometric understanding, while during inference, the full flow $\mathbf{M}_f$ is utilized for final rectification. The rectified image is obtained through differentiable bilinear sampling:
\begin{equation}
\mathbf{I}_r(x, y) = \mathbf{I}_d\big(\mathbf{M}_f(x, y)\big),
\end{equation}
where $\mathbf{M}_f(x, y)$ denotes the predicted sampling coordinates for pixel location $(x, y)$ in the rectified image. All three flows are absolute sampling coordinates rather than displacements.

As illustrated in Fig.~\ref{fig:architecture}, our method consists of five main components: (i) distortion encoding to extract hierarchical features at 1/8 resolution, (ii) dual-branch decoding to capture page-specific deformations, (iii) cross-page attention to model inter-page dependencies, (iv) flow fusion and upsampling to generate high-resolution warping flows, and (v) multi-task training objective.

\subsection{Distortion Encoding}

The distortion encoder of our BookNet extracts hierarchical features through a cascade of CNN~\cite{lecun1998gradient} backbone and Transformer~\cite{Vaswani2017AttentionIA,ViT} encoder, which capture both local geometric patterns and global spatial dependencies. The CNN backbone adopts a lightweight ResNet-style architecture~\cite{he2016deep}, extracting features at 1/8 resolution with channel dimension $C=256$, balancing efficiency and spatial detail preservation.

The extracted features are enhanced through a Transformer encoder with four self-attention layers, eight attention heads, and sinusoidal 2D positional embeddings. The multi-head self-attention mechanism captures long-range dependencies across the book spread, modeling curved text lines and coupled deformation patterns between adjacent pages, which is crucial for understanding bound document geometry. The encoder outputs enhanced features $\mathbf{F}_{enc} \in \mathbb{R}^{\frac{H}{8} \times \frac{W}{8} \times C}$ that encode the book spread geometry, serving as the shared feature representation for subsequent dual-branch decoding.

\subsection{Dual-Branch Decoding}

Each decoder branch employs learnable query embeddings $\mathbf{Q}_l, \mathbf{Q}_r \in \mathbb{R}^{\frac{H}{8} \times \frac{W}{16} \times C}$ that represent spatial regions of the corresponding page. These queries are randomly initialized and optimized during training, serving as rectification anchors where each query is responsible for predicting the warping flow of a specific region in the distorted image.

In the first stage, each decoder branch independently processes its queries through two Transformer decoder layers. The queries attend to the shared encoder features $\mathbf{F}_{enc}$ via multi-head cross-attention to extract page-specific deformation patterns:
\begin{equation}
\begin{aligned}
\mathbf{F}_l^{(1)} &= \mathcal{D}^{(1)}(\mathbf{Q}_l, \mathbf{F}_{enc}), \\
\mathbf{F}_r^{(1)} &= \mathcal{D}^{(1)}(\mathbf{Q}_r, \mathbf{F}_{enc}),
\end{aligned}
\end{equation}
where $\mathcal{D}^{(1)}$ denotes the first-stage decoder (layers 1-2), and $\mathbf{F}_l^{(1)}, \mathbf{F}_r^{(1)} \in \mathbb{R}^{\frac{H}{8} \times \frac{W}{16} \times C}$ capture the local geometric distortions characteristic to each page.

\subsection{Cross-Page Attention}

The second stage introduces bidirectional cross-page attention to enable information exchange between the two branches. We employ encoder-decoder multi-head attention where queries from one page attend to features from the counterpart page:
\begin{equation}
\begin{aligned}
\tilde{\mathbf{F}}_l &= \text{LN}(\mathbf{F}_l^{(1)} + \text{MA}(\mathbf{Q}_l, \mathbf{K}_r, \mathbf{V}_r)), \\
\tilde{\mathbf{F}}_r &= \text{LN}(\mathbf{F}_r^{(1)} + \text{MA}(\mathbf{Q}_r, \mathbf{K}_l, \mathbf{V}_l)),
\end{aligned}
\end{equation}
where $\mathbf{K}_l, \mathbf{V}_l$ and $\mathbf{K}_r, \mathbf{V}_r$ are computed from $\mathbf{F}_l^{(1)}$ and $\mathbf{F}_r^{(1)}$ respectively, $\text{MA}(\cdot, \cdot, \cdot)$ denotes the multi-head attention operation with 8 heads, and $\text{LN}(\cdot)$ represents layer normalization. This cross-page attention mechanism captures spatial correspondences and geometric constraints between adjacent pages, which is crucial for maintaining consistency across the gutter where the two pages meet. The enhanced features $\tilde{\mathbf{F}}_l, \tilde{\mathbf{F}}_r \in \mathbb{R}^{\frac{H}{8} \times \frac{W}{16} \times C}$ then undergo two additional Transformer decoder layers to produce final features:
\begin{equation}
\begin{aligned}
\mathbf{F}_l^{(2)} &= \mathcal{D}^{(2)}(\tilde{\mathbf{F}}_l, \mathbf{F}_{enc}), \\
\mathbf{F}_r^{(2)} &= \mathcal{D}^{(2)}(\tilde{\mathbf{F}}_r, \mathbf{F}_{enc}),
\end{aligned}
\end{equation}
where $\mathcal{D}^{(2)}$ denotes the second-stage decoder (layers 3-4), $\mathbf{F}_l^{(2)}, \mathbf{F}_r^{(2)} \in \mathbb{R}^{\frac{H}{8} \times \frac{W}{16} \times C}$ leverage both page-specific and inter-page information for refined rectification prediction.

\subsection{Flow Fusion and Upsampling}

For comprehensive flow prediction across the book spread, we employ a feature fusion strategy that spatially concatenates the second-stage decoded features $\mathbf{F}_l^{(2)}$ and $\mathbf{F}_r^{(2)}$ along the width dimension to obtain $\mathbf{F}_{concat} \in \mathbb{R}^{\frac{H}{8} \times \frac{W}{8} \times C}$. Two 3$\times$3 convolutional layers with ReLU activation then process the concatenated features to refine the joint representation while preserving spatial correspondence between the two pages.

Each branch generates warping flows through flow prediction heads, which consist of two 3$\times$3 convolutional layers that map features to 2-channel coordinate fields at 1/8 resolution. The coarse coordinate fields are then upsampled to full resolution using a learnable convex upsampling mechanism~\cite{teed2020raft}. For each pixel of the full-resolution output, the network predicts a set of nine softmax-normalized weights over the $3\times3$ neighborhood of the corresponding coarse location, and the fine-resolution value is the weighted average of these nine coarse values. Unlike bilinear interpolation with fixed weights, these weights are predicted from the features, which enables adaptive interpolation that preserves fine document details such as textlines and character boundaries.

The network outputs three warping flows: $\mathbf{M}_l, \mathbf{M}_r \in \mathbb{R}^{H \times \frac{W}{2} \times 2}$ for individual pages, and $\mathbf{M}_f \in \mathbb{R}^{H \times W \times 2}$ for the book spread. Each warping flow defines a backward warping field that samples pixels from the distorted input. This joint supervision strategy leverages both local page-specific deformations and global geometric constraints for robust rectification.

\subsection{Training Objective}

Unlike existing methods that employ complex loss functions with multiple geometric constraints or adversarial training, our approach uses a straightforward multi-task L1 loss that supervises all three warping flows simultaneously:
\begin{equation}
\label{eq:loss}
\mathcal{L} = \|\mathbf{M}_l - \mathbf{M}_l^{gt}\|_1 + \|\mathbf{M}_r - \mathbf{M}_r^{gt}\|_1 + \|\mathbf{M}_f - \mathbf{M}_f^{gt}\|_1,
\end{equation}
where $\mathbf{M}_l^{gt}, \mathbf{M}_r^{gt} \in \mathbb{R}^{H \times \frac{W}{2} \times 2}$, and $\mathbf{M}_f^{gt} \in \mathbb{R}^{H \times W \times 2}$ denote the ground truth warping flows, which record for each pixel of the rectified image the corresponding coordinates in the distorted image. This multi-task supervision enables the network to learn both page-specific deformation patterns and global geometric consistency. By jointly supervising all three outputs with equal weights, the network maintains consistency between individual page rectifications and the book spread, leading to more coherent and visually consistent results across the entire book image. During inference, only $\mathbf{M}_f$ is used for final rectification, benefiting from the comprehensive understanding acquired through multi-task training.

\begin{table}[!t]
\centering
\caption{Quantitative Comparison on Book100 Benchmark}
\label{tab:main_results_book100}
\begin{tabular}{l|ccccc}
\toprule
\textbf{Method} & \textbf{MSSIM} $\uparrow$ & \textbf{LD} $\downarrow$ & \textbf{AD} $\downarrow$ & \textbf{CER} $\downarrow$ & \textbf{ED} $\downarrow$ \\
\midrule
Distorted & 0.28 & 39.63 & 0.96 & 0.0530 & 172.63 \\
\midrule
DewarpNet~\cite{das2019dewarpnet} & 0.41 & 19.98 & 0.62 & 0.0478 & 153.03 \\
DocTr~\cite{feng2021doctr} & 0.47 & 18.10 & 0.59 & 0.0516 & 165.77 \\
DocGeoNet~\cite{feng2022geometric} & \textbf{0.48} & 17.84 & 0.59 & 0.0507 & 166.93 \\
PaperEdge~\cite{ma2022learning} & 0.46 & 14.95 & 0.56 & 0.0472 & 148.10 \\
DocTr++~\cite{feng2023doctrplus} & 0.40 & 28.41 & 0.65 & 0.0559 & 185.43 \\
UVDoc~\cite{UVDoc} & 0.47 & 16.20 & 0.55 & 0.0478 & 166.63 \\
DocRes~\cite{zhang2024docres} & \textbf{0.48} & 16.97 & 0.56 & \textbf{0.0435} & 148.03 \\
\midrule
BookNet (Ours) & \textbf{0.48} & \textbf{12.42} & \textbf{0.53} & 0.0450 & \textbf{141.63} \\
\bottomrule
\end{tabular}
\end{table}

\section{Experiments}
\label{sec:experiments}

\subsection{Implementation Details}

\subsubsection{Training Configuration}
We employ AdamW~\cite{loshchilov2017decoupled} optimizer with OneCycle learning rate schedule, setting maximum learning rate to $1 \times 10^{-4}$ and weight decay to $1 \times 10^{-5}$. The network is trained for 65 epochs with batch size 4 per GPU on 4 NVIDIA RTX 3090 GPUs. Input images are resized to (288, 288). HSV color jittering is applied as data augmentation to enhance robustness to diverse illumination conditions.

\subsubsection{Inference Pipeline}
During the inference stage, distorted images are resized to (288, 288) and produce the left and right flows of (288, 144) and the full flow of (288, 288) for the book spread. The full flow is resized back to the original resolution and applied to the original distorted image via bilinear sampling, so the rectified output keeps the native resolution of the input.

\subsubsection{Evaluation Metrics}
We utilize five metrics for comprehensive evaluation. Specifically, for geometric and perceptual quality assessment, we employ Multi-Scale Structural Similarity (MSSIM)~\cite{1292216,1284395} for perceptual quality, Local Distortion (LD)~\cite{you2018multiview} for geometric accuracy, and Aligned Distortion (AD)~\cite{ma2022learning} for robust distortion measurement. For text recognition accuracy, we use Edit Distance (ED)~\cite{levenshtein1966binary} and Character Error Rate (CER) computed with PaddleOCR-VL~\cite{cui2025paddleocr}. We adopt PaddleOCR-VL rather than traditional OCR engines (\textit{e.g.}, Tesseract) because dual-page book images require accurate separation of left and right page content, which conventional engines fail to handle by mixing text across pages. Following previous work~\cite{UVDoc,zhang2024docres,yang2025adr}, we compute the metrics between the texts recognized from the rectified output and from the reference scan. The same recognizer and configuration are applied to all compared methods. Geometric and perceptual metrics are evaluated on all 100 images, while OCR metrics are computed on 30 representative images with diverse layouts, styles, and content elements. MSSIM weights for the 5 pyramid levels are set as 0.0448, 0.2856, 0.3001, 0.2363, and 0.1333 following previous work~\cite{ma2018docunet,das2019dewarpnet} for fair comparison.

\subsubsection{Model Efficiency}
BookNet is implemented with 30.1M parameters, balancing model capacity and rectification accuracy. The model achieves 30.87 FPS on a single NVIDIA RTX 3090 GPU, demonstrating efficient inference speed for practical document image rectification applications. The FPS is measured at 288$\times$288 with one image at a time, counting the model inference time only.

\begin{figure}[!t]
  \centering
  \includegraphics[width=\linewidth]{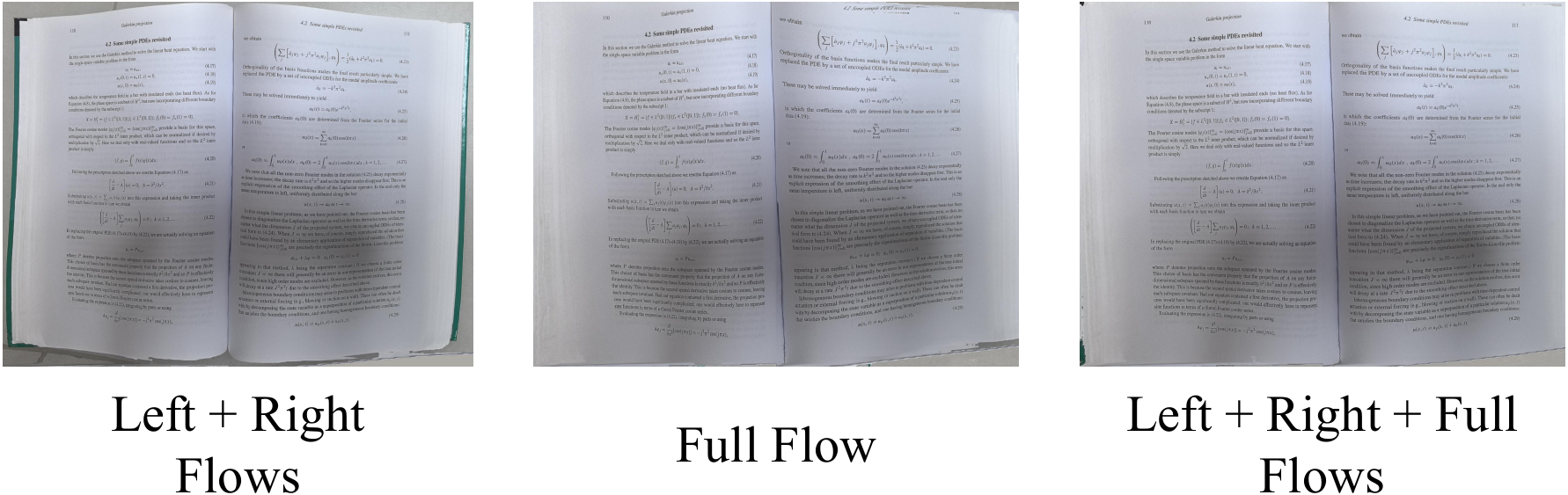}
  \caption{Visual comparison of flow supervision strategies. Left: left and right flows. Middle: full flow only. Right: joint supervision (ours) achieves better gutter alignment.}  
  \label{fig:flow_ablation}
\end{figure}

\subsection{Comparison with State-of-the-Art Methods}

\subsubsection{Quantitative Comparison}
Table~\ref{tab:main_results_book100} presents quantitative comparisons on Book100. BookNet achieves superior performance across most metrics, outperforming the second-best method by 16.9\% on LD and 4.3\% on ED, while matching the highest MSSIM of 0.48 and achieving the best AD of 0.53. These improvements demonstrate that our dual-branch architecture with cross-page attention effectively captures coupled deformation patterns in book pages, leading to accurate geometric rectification with better content preservation and improved text recognition accuracy. The superior performance highlights BookNet's ability to maintain alignment, reduce distortion artifacts, and preserve perceptual quality critical for downstream applications~\cite{wang2011end,lat2018enhancing,peng2022recognition,feng2025dolphin,zhang2022multimodal,kim2022ocr,zhang2025ocrgenbench,chen2026lightweight}.

\begin{figure*}[!t]
 \centering
 \includegraphics[width=1\linewidth]{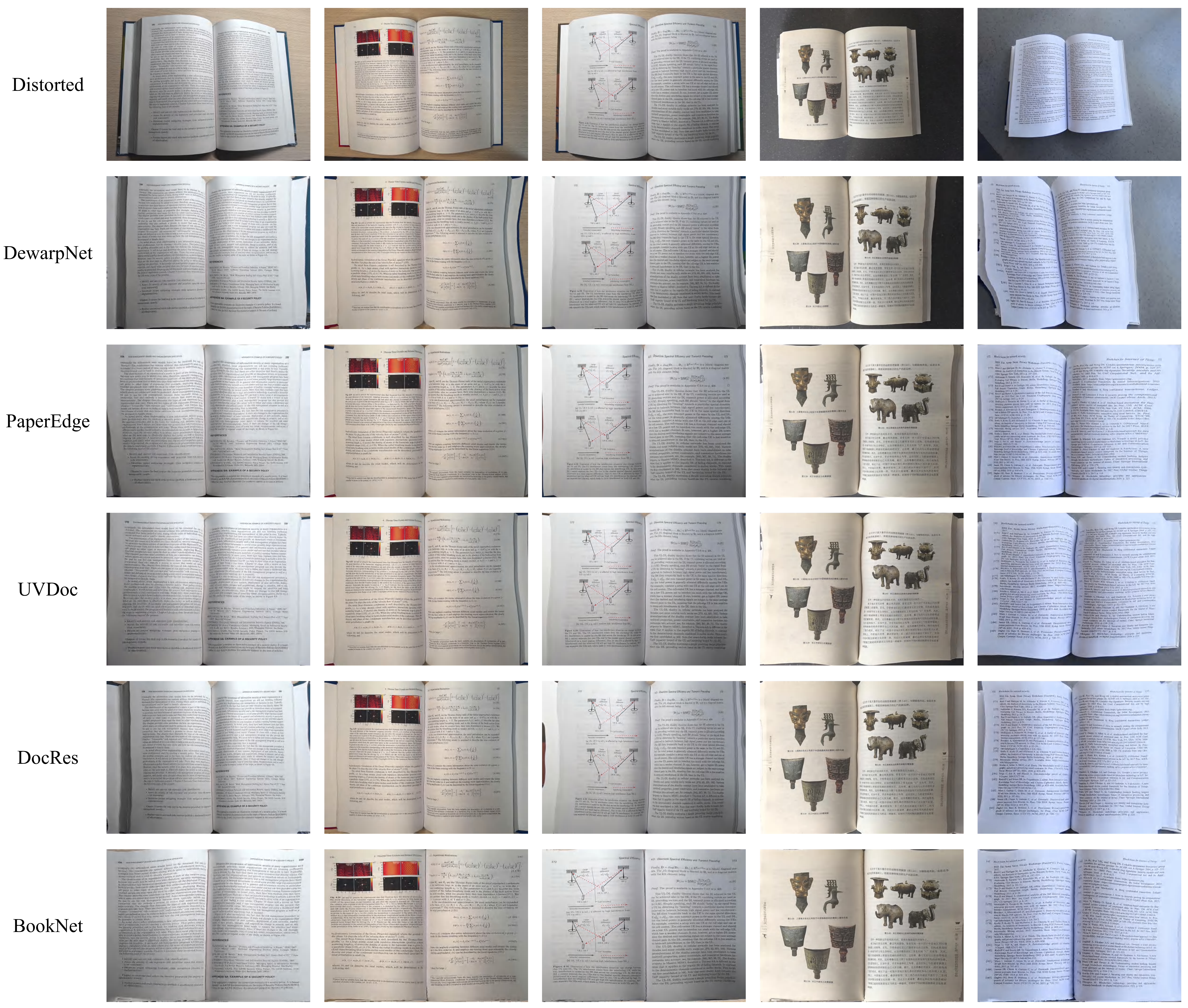}
  \caption{Qualitative comparison of document rectification methods on various book pages. Each row shows results from a different method: Distorted (input), DewarpNet~\cite{das2019dewarpnet}, PaperEdge~\cite{ma2022learning}, UVDoc~\cite{UVDoc}, DocRes~\cite{zhang2024docres}, and BookNet (ours). Our BookNet demonstrates consistent rectification quality across all document types while preserving text readability and image details.}
 \label{fig:qualitative_comparison}
\end{figure*}

\subsubsection{Qualitative Comparison}
Fig.~\ref{fig:qualitative_comparison} shows qualitative comparisons of our BookNet with state-of-the-art methods. Unlike existing single-page methods that often produce inconsistent rectification between pages, BookNet achieves superior geometric consistency across the entire book spread, particularly in the challenging gutter region, while maintaining excellent text line straightening with minimal artifacts. As visible in the results, existing methods struggle with the gutter where the two pages meet, producing visible misalignments or residual curvature, whereas our dual-branch architecture with cross-page attention ensures seamless transitions and consistent rectification quality across both pages. BookNet exhibits remarkable generalization across different document types including pure text, mixed layouts, mathematical formulas, and multiple languages, achieving effective background removal and high-quality page structure reconstruction.

\begin{figure*}[!t]
  \centering
  \includegraphics[width=\linewidth]{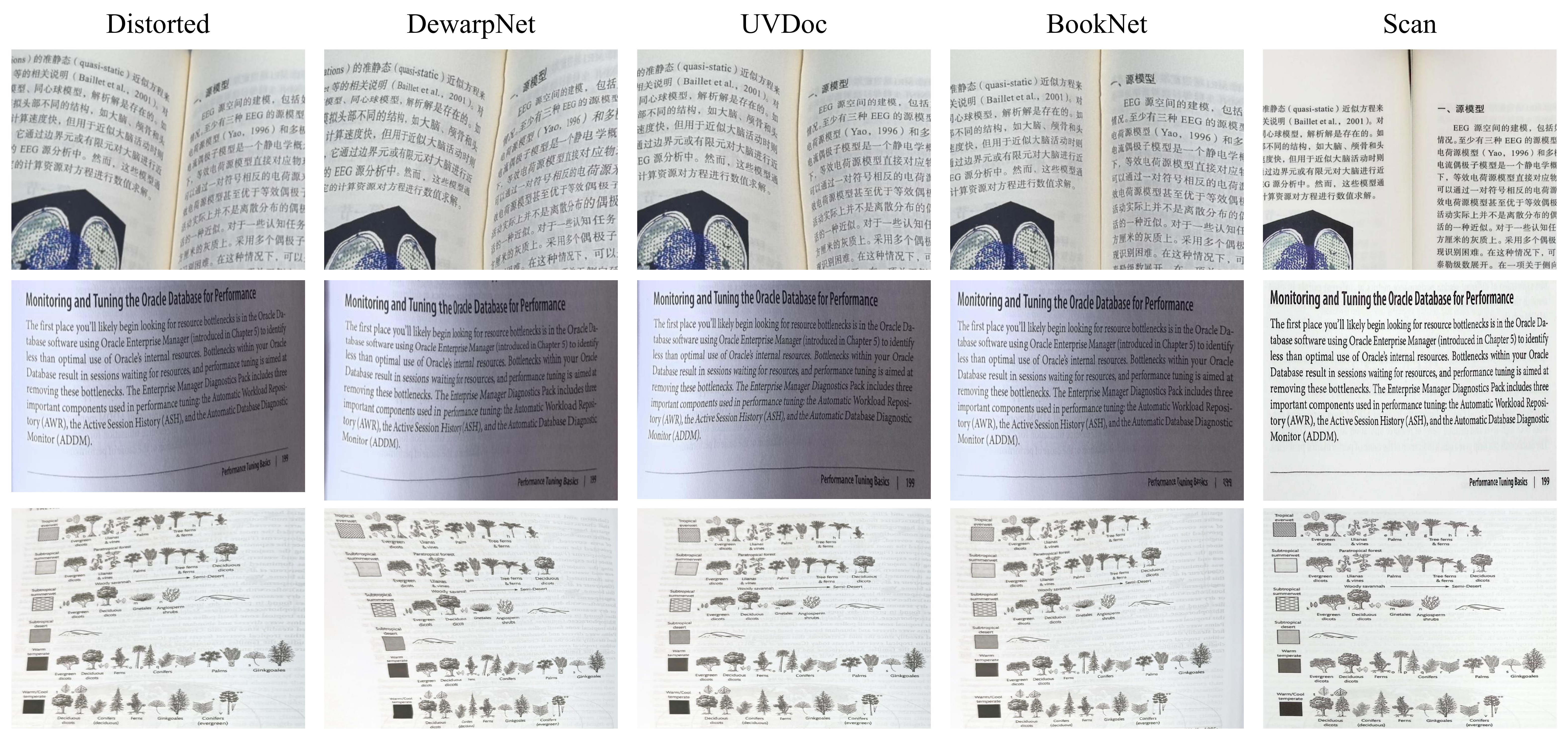}
  \caption{Zoomed-in qualitative comparison. Each example shows the distorted input, the results of DewarpNet~\cite{das2019dewarpnet}, UVDoc~\cite{UVDoc} and BookNet, and the reference scan. The rectified images of BookNet show less distortion than the other rectification methods.}
  \label{fig:qualitative_detail}
\end{figure*}

To allow closer inspection, Fig.~\ref{fig:qualitative_detail} provides zoomed-in crops around the gutter and the text lines. BookNet keeps the text lines straight and the characters legible up to the gutter, while the compared methods leave residual curvature there.

\begin{table}[!t]
\centering
\caption{Ablation Study on Architectural Components}
\label{tab:ablation_architecture}
\setlength{\tabcolsep}{2pt}
\begin{tabular}{ccc|ccccc}
\toprule
\textbf{Trans.} & \textbf{Dual-Br.} & \textbf{Feat. Fusion} & \textbf{MSSIM} $\uparrow$ & \textbf{LD} $\downarrow$ & \textbf{AD} $\downarrow$ & \textbf{CER} $\downarrow$ & \textbf{ED} $\downarrow$ \\
\textbf{Encoder} & \textbf{Decoder} & \textbf{Module} & & & & & \\
\midrule
& \checkmark & \checkmark & 0.47 & 13.75 & 0.55 & 0.0556 & 185.13 \\
\checkmark & & \checkmark & \textbf{0.48} & 12.75 & \textbf{0.52} & 0.0545 & 171.90 \\
\checkmark & \checkmark & & 0.45 & 14.96 & 0.62 & 0.0650 & 206.80 \\
\checkmark & \checkmark & \checkmark & \textbf{0.48} & \textbf{12.42} & 0.53 & \textbf{0.0450} & \textbf{141.63} \\
\bottomrule
\end{tabular}
\end{table}

\begin{table}[!t]
\centering
\caption{Ablation Study on Cross-Page Attention}
\label{tab:ablation_attention}
\setlength{\tabcolsep}{4pt} 
\begin{tabular}{l|ccccc}
\toprule
\textbf{Cross-Page Attention} & \textbf{MSSIM} $\uparrow$ & \textbf{LD} $\downarrow$ & \textbf{AD} $\downarrow$ & \textbf{CER} $\downarrow$ & \textbf{ED} $\downarrow$ \\
\midrule
w/o Cross-Page Attention & 0.46 & 13.11 & 0.58 & 0.0604 & 194.30 \\
w/ Cross-Page Attention & \textbf{0.48} & \textbf{12.42} & \textbf{0.53} & \textbf{0.0450} & \textbf{141.63} \\
\bottomrule
\end{tabular}
\end{table}

\subsection{Ablation Study}

We conduct ablation studies to validate each component in our BookNet architecture and training strategy. All experiments are conducted on Book100 benchmark.

\subsubsection{Flow Supervision Strategy}
Table~\ref{tab:ablation_flow_supervision} compares three supervision approaches. Joint supervision of all three flows achieves the best performance, outperforming page-only supervision by 13.9\% in LD and 30.1\% in ED. As illustrated in Fig.~\ref{fig:flow_ablation}, page-only supervision causes misalignment artifacts at the gutter, as branches independently optimize without considering geometric constraints at the gutter. Full flow supervision maintains global consistency but loses fine-grained page-specific details. Our joint approach combines both strengths: dual page flows capture local deformation patterns while the full flow enforces cross-page geometric consistency.

\begin{table}[!t]
\centering
\caption{Ablation Study on Flow Supervision Strategies}
\label{tab:ablation_flow_supervision}
\setlength{\tabcolsep}{4pt} 
\begin{tabular}{ccc|ccccc}
\toprule
\textbf{Left} & \textbf{Right} & \textbf{Full} & \textbf{MSSIM} $\uparrow$ & \textbf{LD} $\downarrow$ & \textbf{AD} $\downarrow$ & \textbf{CER} $\downarrow$ & \textbf{ED} $\downarrow$ \\
\midrule
\checkmark & \checkmark & & 0.47 & 14.43 & 0.60 & 0.0611 & 202.57 \\
& & \checkmark & 0.45 & 15.01 & 0.61 & 0.0566 & 186.13 \\
\checkmark & \checkmark & \checkmark & \textbf{0.48} & \textbf{12.42} & \textbf{0.53} & \textbf{0.0450} & \textbf{141.63} \\
\bottomrule
\end{tabular}
\end{table}

\begin{table}[!t]
\centering
\caption{Ablation Study on Loss Weights}
\label{tab:ablation_loss_weight}
\setlength{\tabcolsep}{10pt}
\begin{tabular}{c|ccc}
\toprule
$\boldsymbol{(\lambda_l, \lambda_r, \lambda_f)}$ & \textbf{MSSIM} $\uparrow$ & \textbf{LD} $\downarrow$ & \textbf{AD} $\downarrow$ \\
\midrule
$(2,2,1)$ & 0.47 & 12.75 & 0.58 \\
$(1,1,2)$ & 0.47 & 12.91 & 0.57 \\
$(1,1,1)$ & \textbf{0.48} & \textbf{12.42} & \textbf{0.53} \\
\bottomrule
\end{tabular}
\end{table}

\subsubsection{Cross-Page Attention Mechanism}
Table~\ref{tab:ablation_attention} demonstrates that cross-page attention provides substantial improvements, achieving 5.3\% reduction in LD, 25.5\% reduction in CER and 27.1\% reduction in ED. Without this mechanism, dual branches process pages independently, failing to capture the coupled deformation patterns fundamental to book geometry. Cross-page attention enables bidirectional information exchange between branches, maintaining both geometric and semantic consistency across the book spread and consistently improving text recognition accuracy by ensuring coherent rectification. This mechanism is particularly effective for handling the asymmetric distortions in book pages where left and right pages influence each other through binding constraints.

\begin{figure}[!t]
  \centering
  \includegraphics[width=\linewidth]{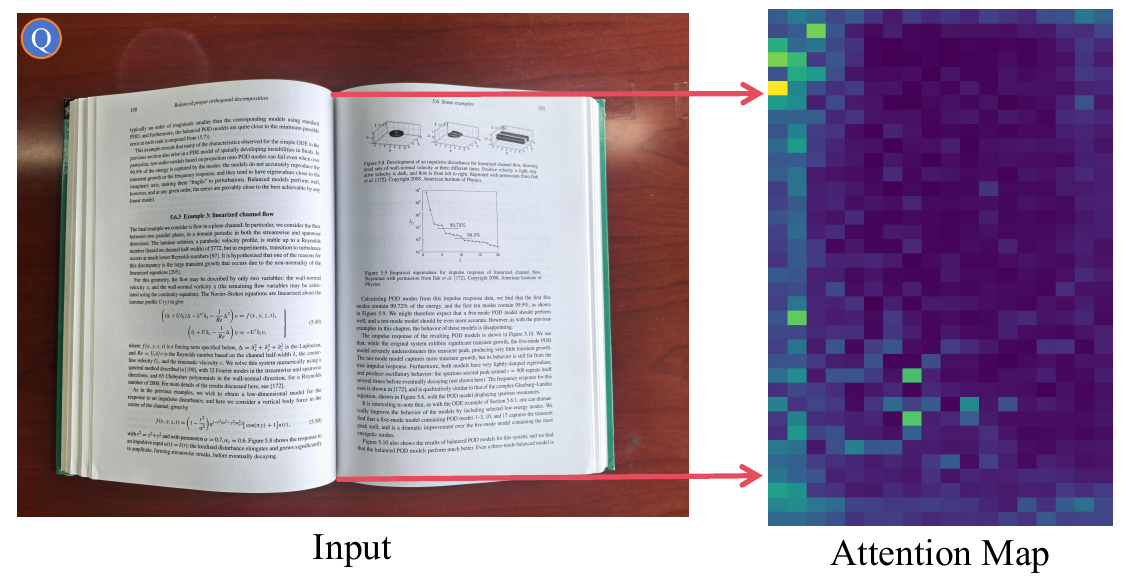}
  \caption{Visualization of cross-page attention. The query point at the upper-left corner of the input image attends to the gutter region of the right page, demonstrating learned geometric coupling across pages.}
  \label{fig:attention_vis}
\end{figure}

To further understand how cross-page attention operates, we visualize the attention maps in Fig.~\ref{fig:attention_vis}. Given a query position at the upper-left corner of the input image, we observe that the cross-page attention predominantly attends to the corresponding gutter region of the right page. This confirms that our mechanism learns meaningful geometric coupling between adjacent pages rather than performing generic feature mixing, enabling precise alignment at the gutter where binding-induced distortions are most severe.

\begin{figure*}[!t]
  \centering
  \includegraphics[width=0.9\linewidth]{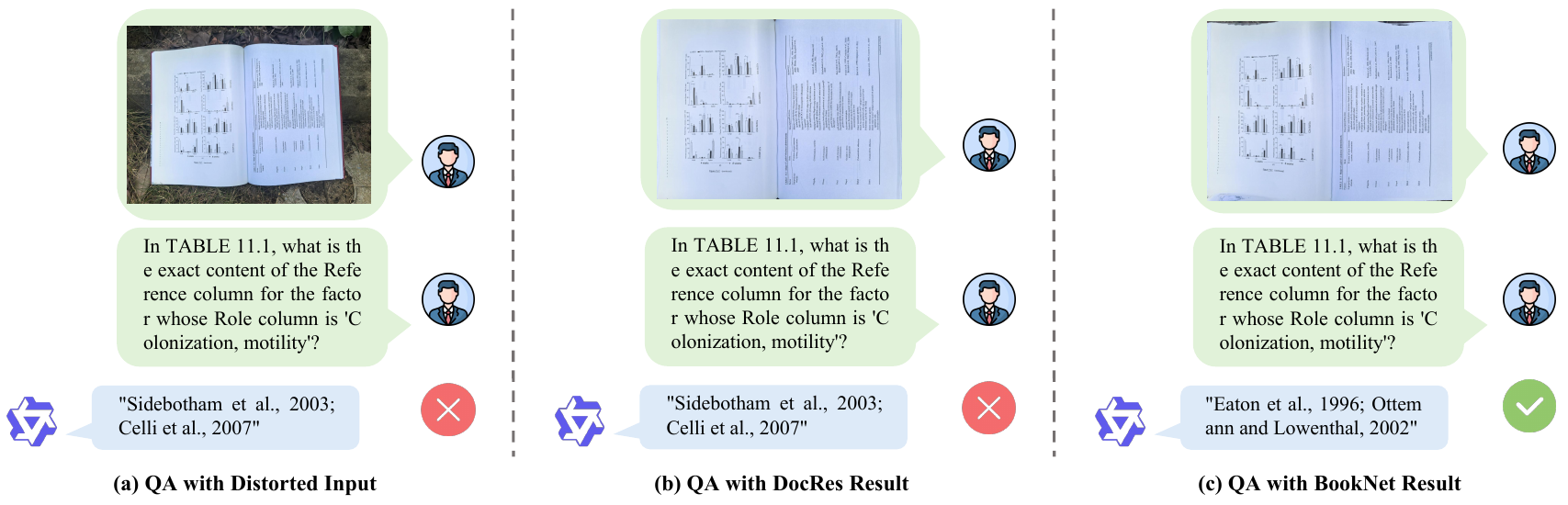}
  \caption{Representative VQA case on the distorted input, the DocRes result, and the BookNet result, evaluated with Qwen2.5-VL-7B.}
  \label{fig:vqa_cases}
\end{figure*}

\begin{figure}[!t]
  \centering
  \includegraphics[width=\linewidth]{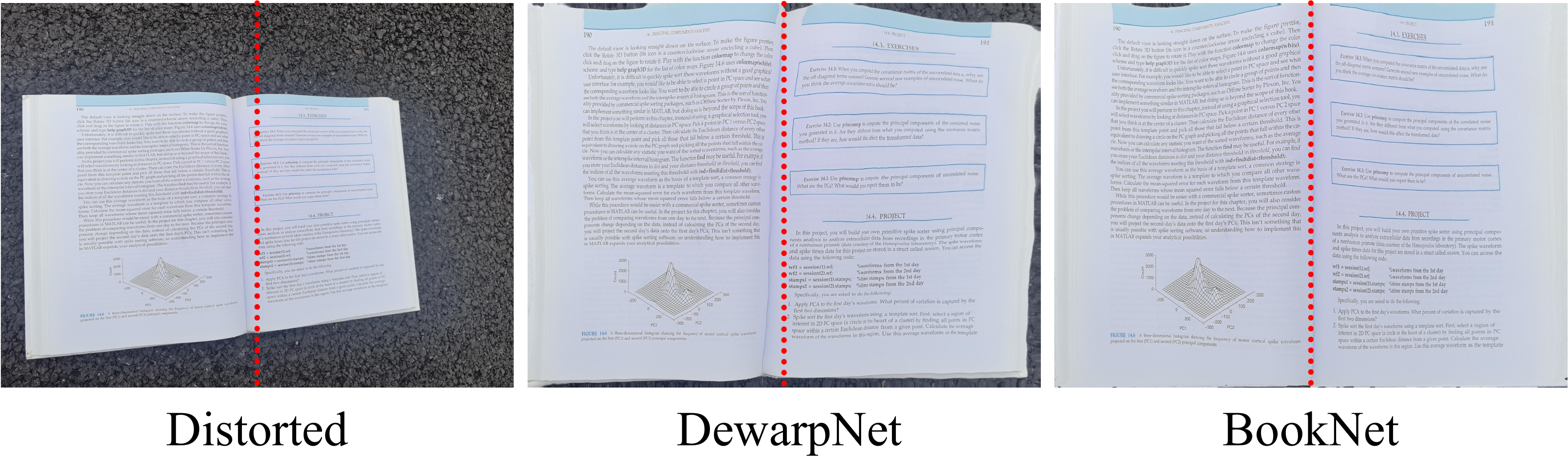}
  \caption{Robustness to off-center layouts. BookNet produces accurate rectification even when the book spine is significantly off-center.}
  \label{fig:offcenter}
\end{figure}

\subsubsection{Architecture Component Analysis}
Table~\ref{tab:ablation_architecture} examines key components. Removing the feature fusion module causes the largest degradation across all metrics. Without it, LD rises from 12.42 to 14.96 and ED rises from 141.63 to 206.80, which shows that a joint representation over the two pages is necessary before flow prediction. Removing the Transformer encoder also degrades text recognition, increasing CER from 0.0450 to 0.0556, which confirms its role in capturing long-range dependencies crucial for text readability. The dual-branch decoder contributes as well. Their synergistic combination demonstrates each module is indispensable.

\subsubsection{Loss Weight Sensitivity}
The three flows play different roles, so we further examine whether equal weighting is a reasonable choice. We generalize the training objective in Eq.~\eqref{eq:loss} to
\begin{equation}
\mathcal{L} = \lambda_l \|\mathbf{M}_l - \mathbf{M}_l^{gt}\|_1 + \lambda_r \|\mathbf{M}_r - \mathbf{M}_r^{gt}\|_1 + \lambda_f \|\mathbf{M}_f - \mathbf{M}_f^{gt}\|_1 ,
\end{equation}
where we keep $\lambda_l = \lambda_r$, since the left and right pages play symmetric roles. Table~\ref{tab:ablation_loss_weight} compares three settings of the weight triplet $(\lambda_l, \lambda_r, \lambda_f)$. Besides the equal weighting $(1,1,1)$ adopted in our final model, we consider two unbalanced settings: $(2,2,1)$, which weights each page-specific flow twice as much as the full flow, and $(1,1,2)$, which weights the full flow twice as much as each page-specific flow. Both unbalanced settings perform worse than equal weighting. Equal weighting achieves an MSSIM of 0.48 and an LD of 12.42, whereas $(2,2,1)$ and $(1,1,2)$ yield a lower MSSIM of 0.47 and higher LD values of 12.75 and 12.91, respectively. We therefore retain equal weighting in our final model.

\subsubsection{Input Resolution}
We further train our model at 384$\times$384 and 512$\times$512 to examine the trade-off between accuracy and efficiency. As shown in Table~\ref{tab:ablation_resolution}, increasing the input resolution does not bring a consistent gain in accuracy. The 384$\times$384 model gives an LD of 12.43, essentially the same as the 12.42 obtained at 288$\times$288, while MSSIM decreases from 0.48 to 0.47 and AD increases from 0.53 to 0.56. The 512$\times$512 model reduces LD to 12.29, but MSSIM and AD remain at 0.47 and 0.56. Meanwhile, the inference speed drops from 30.87 FPS at 288$\times$288 to 25.54 FPS at 384$\times$384, and further to 10.40 FPS at 512$\times$512, roughly one third of the original speed. Since an LD reduction of 0.13 does not justify such an increase in inference cost, we adopt 288$\times$288 as the default input resolution for practical deployment.

\begin{table}[!t]
\centering
\caption{Ablation Study on Input Resolution}
\label{tab:ablation_resolution}
\setlength{\tabcolsep}{10pt}
\begin{tabular}{c|ccc|c}
\toprule
\textbf{Resolution} & \textbf{MSSIM} $\uparrow$ & \textbf{LD} $\downarrow$ & \textbf{AD} $\downarrow$ & \textbf{FPS} $\uparrow$ \\
\midrule
288$\times$288 & \textbf{0.48} & 12.42 & \textbf{0.53} & \textbf{30.87} \\
384$\times$384 & 0.47 & 12.43 & 0.56 & 25.54 \\
512$\times$512 & 0.47 & \textbf{12.29} & 0.56 & 10.40 \\
\bottomrule
\end{tabular}
\end{table}

\begin{table}[!t]
\centering
\caption{Quantitative VQA Evaluation}
\label{tab:vqa}
\setlength{\tabcolsep}{4pt}
\begin{tabular}{l|ccc}
\toprule
\textbf{Version} & \textbf{Distorted} & \textbf{DocRes} & \textbf{BookNet (Ours)} \\
\midrule
\textbf{Accuracy} & 20/30 (66.7\%) & 25/30 (83.3\%) & \textbf{28/30 (93.3\%)} \\
\bottomrule
\end{tabular}
\end{table}

\subsubsection{Robustness to Off-Center Layouts}
A practical concern is whether BookNet requires the book spine to be centered in the input image. We evaluate our method on off-center captures where the spine deviates significantly from the image center due to tilted viewing angles. As shown in Fig.~\ref{fig:offcenter}, BookNet successfully handles such cases, producing well-rectified results even when the spine is off-center, while existing methods struggle in such cases. This robustness is attributed to our diverse training viewpoints in Book3D and the cross-page attention mechanism, which adaptively models inter-page relationships without relying on explicit spatial priors about spine location.

\subsubsection{Impact on Downstream Tasks}
We evaluate BookNet as preprocessing for multimodal understanding using Qwen2.5-VL-7B~\cite{bai2025qwen2}. We build 30 question-answer pairs over Book100 from the content of the reference scans, and query Qwen2.5-VL-7B with the same prompts on three versions of each image. The question-answer pairs are generated with GLM-5.3-Flash and then manually reviewed. The predicted answers are judged against the ground truth by GLM-5.3-Flash, and every judgment is manually verified. As reported in Table~\ref{tab:vqa} and Fig.~\ref{fig:vqa_cases}, the model answers 28 out of 30 correctly on the BookNet results, ahead of DocRes and the distorted input, showing a clear benefit for multimodal understanding.

\section{Limitation Discussion}
First, our BookNet is designed around joint supervision of the left, right, and full flows. Therefore, the current model cannot be directly applied to single-page documents. Second, a unified model that handles both single-page documents and books is a meaningful research problem, and it is of clear practical value, since the input type is often unknown in advance. We will explore this direction in our future work.

\section{Conclusion}
\label{sec:conclusion}

In this paper, we present BookNet, the first end-to-end deep learning framework for dual-page book image rectification. Our dual-branch architecture with cross-page attention simultaneously processes left and right pages while explicitly modeling their geometric interdependencies. To facilitate book image rectification research, we contribute comprehensive datasets (Book3D and Book100) and establish evaluation protocols specifically tailored for this task. Extensive experiments demonstrate that BookNet significantly outperforms state-of-the-art methods, achieving superior performance across multiple metrics on the Book100 benchmark. We hope that our method can serve as a strong baseline for the community and benefit downstream document analysis, understanding, and multimodal applications.

\bibliographystyle{IEEEtran}
\bibliography{main}

\clearpage

\end{document}